\documentclass[lettersize,journal]{IEEEtran}
\usepackage{newtxtext}   % Times 文本字体（支持斜体、粗体）
\usepackage{newtxmath}   % Times 数学字体（可选）
\usepackage{fontspec}   % For XeLaTeX font handling
\usepackage{graphicx} \graphicspath{ {Figures/} }
\usepackage{mathtools}
\usepackage{algorithmic}
\usepackage{mathrsfs}
\usepackage[ruled]{algorithm2e}
\usepackage{acronym}
\usepackage{enumitem}
\usepackage{booktabs}
\usepackage[colorlinks,linkcolor=blue,anchorcolor=green,citecolor=blue,urlcolor=blue]{hyperref}
\usepackage[caption=false,font=normalsize,labelfont=sf,textfont=sf]{subfig}
\usepackage{flushend}
\usepackage{balance}
\usepackage{xspace,setspace}
\usepackage[dvipsnames]{xcolor}
\usepackage[capitalise]{cleveref}
\usepackage{tabularx,colortbl,multirow,array,makecell}
\usepackage{overpic}
\usepackage{tikz}

\makeatletter
\DeclareRobustCommand\onedot{\futurelet\@let@token\@onedot}
\def\@onedot{\ifx\@let@token.\else.\null\fi\xspace}

\makeatother

\crefname{algocf}{alg.}{algs.}
\Crefname{algocf}{Algorithm}{Algorithms}

\crefname{algocf}{alg.}{algs.}
\Crefname{algocf}{Algorithm}{Algorithms}

\definecolor{gblue}{HTML}{4285F4}
\definecolor{gred}{HTML}{DB4437}

\usepackage{textcomp}

\usepackage{tabularx,colortbl,multirow,array,makecell}
\usepackage{threeparttable} % 增加表注
\usepackage[marginal]{footmisc}

\usepackage{makecell}

\usepackage{soul, color}
\sethlcolor{yellow} %pink, blue 等
\soulregister{\cite}7 %当highlight中又引用的话，也一起标黄
\soulregister{\ref}7
\usepackage[backend=biber,
            style=numeric-comp,
            sorting=none,          % 按引用顺序编号
            giveninits=true,       % 作者名缩写（例如 D. Rus）
            maxbibnames=99,        % 显示所有作者（超出用 et al.）
            minbibnames=3,
            doi=true,             % 我们手动控制 DOI 输出
            url=false,
            isbn=false,
            eprint=false]{biblatex}
\usepackage{xurl}                % 使 DOI 长链接可以自动换行

\DeclareFieldFormat{labelnumberwidth}{#1\adddot\addspace}
\defbibenvironment{bibliography}
  {\list
     {\printtext[labelnumberwidth]{%
        \printfield{labelnumber}%
        \adddot\addspace}}
     {\setlength{\labelwidth}{\labelnumberwidth}%
      \setlength{\leftmargin}{\labelwidth}%
      \setlength{\labelsep}{\biblabelsep}%
      \addtolength{\leftmargin}{\labelsep}%
      \setlength{\itemsep}{\bibitemsep}%
      \setlength{\parsep}{\bibparsep}}}
  {\endlist}
  {\item}

\DeclareNameAlias{author}{given-family}
\DeclareNameAlias{editor}{given-family}
\DeclareNameAlias{translator}{given-family}

\DeclareFieldFormat{title}{\mkbibquote{#1\isdot}}          % 双引号
\DeclareFieldFormat{booktitle}{\mkbibemph{#1}}             % 书名斜体
\DeclareFieldFormat{journaltitle}{\mkbibemph{#1}}          % 期刊名斜体
\DeclareFieldFormat{journal}{\mkbibemph{#1}}               % 期刊名斜体
\DeclareFieldFormat{pages}{#1}                             % 不加 “pp.”
\renewbibmacro*{in:}{}                                     % 去掉 “In:”

\renewbibmacro*{volume+number+eid}{%
  \printfield{volume}%
  \setunit*{\addcomma\space}%
  \printfield{number}%
  \setunit{\addcomma\space}%
  \printfield{pages}%
}
\renewbibmacro*{journal+issuetitle}{%
  \usebibmacro{journal}%
  \setunit*{\addspace}%
  \printtext[parens]{\usebibmacro{date}}%  年份用括号
  \setunit*{\addcomma\space}%
  \usebibmacro{volume+number+eid}%
}

\renewbibmacro*{date}{%
  \printfield{year}%
}

\DeclareFieldFormat{doi}{%
  \href{https://doi.org/#1}{#1}%
}
\newbibmacro*{mydoi}{%
  \iffieldundef{doi}{}%
    {\addcomma\space\printfield{doi}}%
}

\DeclareBibliographyDriver{article}{%
  \usebibmacro{bibindex}%
  \usebibmacro{begentry}%
  \usebibmacro{author/editor}%
  \setunit{\addcomma\space}%
  \printfield{title}%
  \setunit{\addcomma\space}%
  \usebibmacro{journal}% 
  \setunit*{\addspace}%
  \printfield{volume}%                     卷号
  \setunit*{\addcomma\space}%
  \iffieldundef{number}{}%                ← 条件判断：无期号则不输出 "no."
    {\printtext{no.\addspace}\printfield{number}}%
  \setunit*{\space}%
  \printtext[parens]{\usebibmacro{date}}%  (年份)
  \setunit*{\addcolon\addspace}%           : 空格
  \printfield{pages}%                      页码
  \newunit
  \usebibmacro{mydoi}%                     DOI
  \usebibmacro{finentry}%
}

\DeclareBibliographyDriver{book}{%
  \usebibmacro{bibindex}%
  \usebibmacro{begentry}%
  \usebibmacro{author}%
  \setunit{\addperiod\addspace}%
  \mkbibemph{\printfield{title}}%
  \setunit{\addperiod\addspace}%
  \printfield{edition}%
  \setunit{\addperiod\addspace}%
  (%
    \printlist{publisher}%
    \setunit*{\addcomma\space}%
    \printlist{location}%
    \setunit*{\addcomma\space}%
    \printfield{year}%
  )%
  \setunit*{\addcomma\addspace}%
  \printfield{pages}%
  \newunit%
  \usebibmacro{mydoi}%
  \usebibmacro{finentry}%
}

\DeclareBibliographyDriver{inbook}{%
  \usebibmacro{bibindex}%
  \usebibmacro{begentry}%
  \usebibmacro{author/editor}%
  \setunit{\addcomma\space}%
  \printfield{title}%
  \setunit{\addcomma\space}%
  \printfield{booktitle}%
  \setunit*{\addcomma\space}%
  \printfield{volume}%
  \setunit*{\addcomma\space}%
  \printfield{pages}%
  \setunit{\addspace}%
  \printtext[parens]{\usebibmacro{date}}%
  \newunit
  \usebibmacro{mydoi}%
  \usebibmacro{finentry}%
}
\DeclareBibliographyAlias{incollection}{inbook}   % 让 incollection 同 inbook

\DeclareBibliographyDriver{inproceedings}{%
  \usebibmacro{bibindex}%
  \usebibmacro{begentry}%
  \usebibmacro{author/editor}%
  \setunit{\addcomma\space}%
  \printfield{title}%
  \setunit{\addcomma\space}%
  \printfield{booktitle}%     会议论文集名
  \setunit*{\addcomma\space}%
  \printfield{pages}%
  \setunit{\addspace}%
  \printtext[parens]{\usebibmacro{date}}%
  \newunit
  \usebibmacro{mydoi}%
  \usebibmacro{finentry}%
}

\DeclareBibliographyDriver{proceedings}{%
  \usebibmacro{bibindex}%
  \usebibmacro{begentry}%
  \usebibmacro{editor}%
  \setunit{\addcomma\space}%
  \printfield{booktitle}%
  \setunit{\addspace}%
  \printtext[parens]{\usebibmacro{date}}%
  \newunit
  \usebibmacro{mydoi}%
  \usebibmacro{finentry}%
}

\AtEveryBibitem{\clearfield{url}}

\makeatletter
\Gin@draftfalse
\makeatother

\begin{document}
\title{Improving Human Diving Endurance with a Field-Deployable Exoskeleton}

\author{Zhihao Zhou,
        Zhenmeng Ju,
        Chenxi Zhang,
        Ming Xu,
        Enhao Zheng,
        Lecheng Ruan,
        and Qining Wang
       % <-this % stops a space

\thanks{This work was supported by the National Natural Science Foundation of China (No. 52375001, No. 52475001, and No. 524B2044). 
\emph{(Zhihao Zhou and Zhenmeng Ju contributed equally to this work. Corresponding author: Qining Wang)}}
\thanks{This work involved human subjects or animals in its research. The approval of all ethical and experimental procedures and protocols was granted by the Ethical Review Board of the Institute of Automation, Chinese Academy of Sciences.}
\thanks{Zhihao Zhou is with the Institute for Artificial Intelligence, Peking University, Beijing 100871, China. (e-mail: zhihaozhou@pku.edu.cn).}
\thanks{Zhengmeng Ju, Chenxi Zhang, Ming Xu, and Lecheng Ruan are with the School of Advanced Manufacturing and Robotics, Peking University, Beijing 100871, China.}
\thanks{Enhao Zheng is with the State Key Laboratory of Multimodal Artificial Intelligence Systems, Institute of Automation, Chinese Academy of Sciences, Beijing, China.}
\thanks{Qining Wang is with the School of Advanced Manufacturing and Robotics, and the Institute for Artificial Intelligence, Peking University, Beijing 100871, China. (e-mail: qiningwang@pku.edu.cn).}
}

\maketitle
\begin{abstract}
Human endurance in underwater locomotion is fundamentally restricted by high energetic demands to overcome drag while managing a finite supply of self-contained breathing gas. %Methods for improving human diving endurance in realistic underwater environments remain scarce. 
Here, we present DiveMate, a field-deployable exoskeleton designed to improve human diving endurance via kick assistance in real-world underwater environments. DiveMate increases the travel distance using a given amount of breathing gas by 42.9\% and extends dive duration by 54.9\% through reducing the gas consumption rate. Marked reductions in muscle activation indicate decreased physiological exertion, with the net gas consumption rate decreasing by 47.0\%. Changes in kinematic characteristics further contribute to powerful and efficient kicking. Successful open-water trials at depths up to 30 m demonstrated field feasibility. These results suggest that applying exoskeleton assistance is beneficial for improving human diving endurance and augmenting their ability to explore the aquatic world. This study extends the application frontier of exoskeletons and provides a potential reference for the design and assessment of future underwater assistive devices.
\end{abstract}

% \twocolumn[%
%   \begin{@twocolumnfalse}
%     \maketitle
%     \begin{abstract}
%     \normalsize
%     Human endurance in underwater locomotion is fundamentally restricted by high energetic demands to overcome drag while managing a finite supply of self-contained breathing gas. %Methods for improving human diving endurance in realistic underwater environments remain scarce. 
%     Here, we present DiveMate, a field-deployable exoskeleton designed to improve human diving endurance via kick assistance in real-world underwater environments. DiveMate increases the travel distance using a given amount of breathing gas by 42.9\% and extends dive duration by 54.9\% through reducing the gas consumption rate. Marked reductions in muscle activation indicate decreased physiological exertion, with the net gas consumption rate decreasing by 47.0\%. Changes in kinematic characteristics further contribute to powerful and efficient kicking. Successful open-water trials at depths up to 30 m demonstrated field feasibility. These results suggest that applying exoskeleton assistance is beneficial for improving human diving endurance and augmenting their ability to explore the aquatic world. This study extends the application frontier of exoskeletons and provides a potential reference for the design and assessment of future underwater assistive devices.
%     \end{abstract}
%     % \begin{IEEEkeywords}
%     % Human endurance, efficient kicking, exoskeleton technology, field-deployable
%     % \end{IEEEkeywords}
%     \vspace{\baselineskip} 
%     \vspace{\baselineskip} 
%   \end{@twocolumnfalse}]

\section{Introduction}
Numerous high-stakes diving activities that rely on divers' essential operations impose extreme physiological demands on human endurance, such as aquaculture harvesting, offshore extraction, emergency rescue, and marine exploration \cite{taylor1957, Koreafemale2025, DivingManual2018}. As humans adapt to the terrestrial locomotor environment evolutionarily \cite{Sockol2007, Forssberg1985}, they become exceptional walkers \cite{Collins2015Nature} and exhibit inefficient swimming locomotion compared to marine animals (e.g., fish) \cite{Goff1957JAP, Pendergast2015CP}, which are morphologically and physiologically adapted to the aquatic environment. During human locomotion, the resistance that the body encounters in water is much greater than that in air, demanding substantial muscular effort and breath \cite{Pendergast2003JAP, TianZeng2015ABB, Grimshaw2007NOTES}. Humans who rely on pulmonary respiration must carry a self-contained air supply for prolonged submerged sustainment \cite{Pendergast1996MSSE, Lambert2024EMC}. This physiological burden is further restricted by the limited capacity of a high-pressure tank carried by divers (e.g., a typical 12 L cylinder pressurized to 20 MPa). These combined factors severely restrict the underwater duration and transport range of divers. Therefore, improving human endurance is imperative for traveling farther and staying longer to complete underwater work \cite{Pendergast2015CP}, thereby demanding innovative solutions to overcome physiological challenges in aquatic environments.

\begin{figure*}[htbp]
    \includegraphics[width=0.95\textwidth]{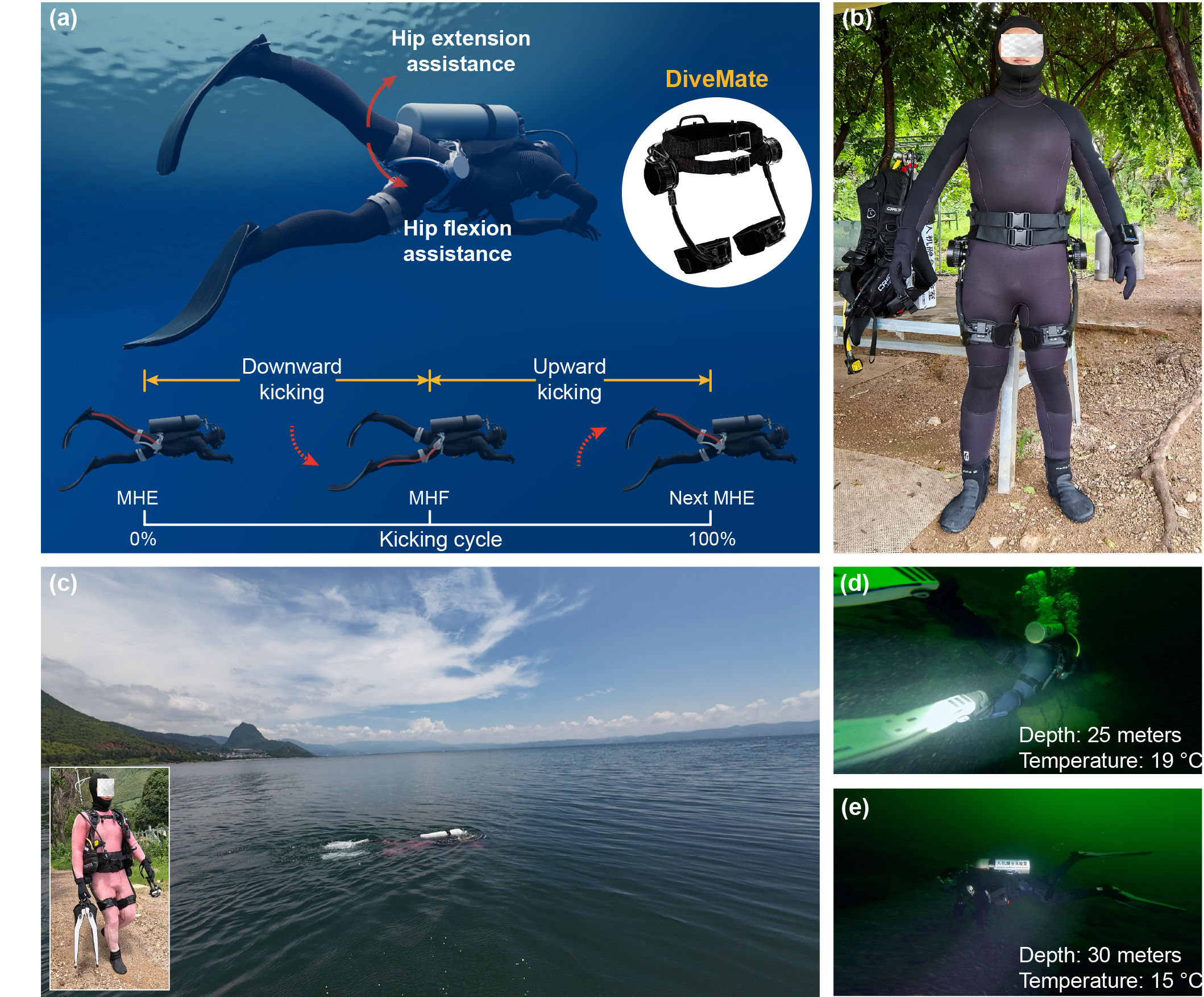}
    \caption{DiveMate exoskeleton design for diving endurance improvement. (a) The untethered exoskeleton for human kicking assistance and endurance improvement in underwater diving scenarios. The kicking cycle (right leg exemplar) initiates at maximum hip extension (MHE). Each cycle comprises two temporally partitioned phases: downward kicking and upward kicking, demarcated by maximum hip flexion (MHF). (b) A subject dons the proposed DiveMate. (c) DiveMate can be deployed in open-water environments and seamlessly integrates with standard scuba equipment, making it an unobtrusive assistive tool for daily diving. (d and e)  An open-water trial was conducted at Fuxian Lake (a surface area of 216 square kilometers, an average depth of 95 m, and a maximum depth of 158 m). The first and second authors dived to a depth of 30 meters in the lake to verify the feasibility of the DiveMate.}
    \label{Fig1}
\end{figure*}

Despite the importance of endurance in diving, research on its multifactorial limiting mechanisms remains lacking. During diving, endurance is shaped by the interplay of several physiological and biomechanical variables, including respiratory gas consumption, muscular activation patterns, and overall energy expenditure. Among these, respiratory muscle fatigue has become a particularly well-documented limiting factor \cite{Pendergast2007EJAP}. The immersive aquatic environment imposes hydrostatic pressure differences across the chest wall, substantially increasing the mechanical work required for inspiration. Under prolonged submersion, this elevated inspiratory load accelerates the onset of respiratory muscle fatigue, which in turn compromises ventilatory efficiency and curtails sustainable swimming duration. Enhancing the diver's intrinsic physiological capacity is one approach to the endurance problem; an equally important yet comparatively underexplored strategy is to reduce the physiological cost imposed by diving itself. By decreasing gas consumption and lowering muscle activation during fin-kicking locomotion, it is possible to improve the overall energy economy of diving.

The concept of underwater exoskeletons has been proposed \cite{Neuhaus2004ICRA} and explored in prior work, with prototype systems developed for both general aquatic locomotion \cite{WangQN2020TMRB, ChenLW2024JFR} and SCUBA diving applications \cite{WuXD2025TRO, WuXY2025TRO}. While these studies demonstrated that powered joint assistance can reduce the energetic cost of diving, their scope remains limited in several critical respects. First, existing work has focused narrowly on reductions in energy expenditure, without directly assessing diving endurance, yet endurance is one of the best indicators of a diver's performance. Second, none have implemented continuous real-time physiological monitoring throughout the diving trial, precluding characterization of the temporal dynamics of gas consumption. Third, all existing evaluations were conducted in controlled laboratory settings, such as water tanks or swimming pools. Validation has never been performed in open water. Beyond these gaps, to be truly practical for real-world diving operations, an underwater exoskeleton must meet three requirements: compatibility with standard diving equipment, the ability to adapt to moving-to-stationary transitions and variable kicking pace, and field deployability. To date, no system has been found to meet all of these requirements simultaneously in real-world applications.

Motivated by the gap identified above, this paper presents the design and experimental evaluation of a field-deployable underwater exoskeleton, hereafter referred to as DiveMate (as depicted in Fig. \ref{Fig1}), intended for use in open-water diving environments (detailed in Supplementary Videos \textcolor{blue}{1} and \textcolor{blue}{2}). By applying an assistive torque to the hip joint during the downward and upward phases of the flutter kick, the exoskeleton provides kinetic assistance to the diver’s kicking motion, thereby offloading a portion of the mechanical work normally borne by the lower limb muscle groups. By reducing muscle activation and respiratory load during continuous kicking, the system aims to lower gas consumption and overall physiological cost, ultimately improving the energy efficiency of underwater locomotion and enhancing divers' underwater endurance.

This study contributes twofold to the field of aquatic augmentation technology. First, we propose a field-deployed exoskeleton design for real-world applications that fully accounts for mechanical hardware and motion-adaptive control. Second, we conduct a systematic experimental study to quantify the multi-dimensional physiological and biomechanical effects of exoskeleton assistance on diving endurance. Specifically, we evaluate the impact of powered assistance across key metrics: respiratory gas consumption rate, lower-limb muscle activation, sustainable diving time, and travel distance. By characterizing these variables, this study offers a multi-dimensional assessment of how assistive underwater exoskeletons affect human diving endurance, with implications for the future design and real-world deployment of such systems.
\section{Exoskeleton Design}
\subsection{Biological Requirements}
The flutter kick is the most typical propulsion skill in diving \cite{Pendergast1996MSSE, Samimy2005SportsEng}. Humans perform alternating upward and downward movements of both legs with fins \cite{Sanders2018JSS, Pendergast2003UHMed}. The downward kick is initiated by hip flexion while maintaining knee extension and ankle dorsiflexion. At the same time, the fin moves vertically with respect to the water surface to generate thrust \cite{Marion2010PE}. The upward kick is achieved by hip extension to lift the lower limb in preparation for the next downward kick \cite{Zamparo2002JEB, Wojtkow2017ABB}. Underwater propulsion requires greater muscle activation during hip flexion and extension than level-ground walking. This may result in substantial gas consumption and potential muscle fatigue \cite{Samimy2005SportsEng}. Since the hip plays a crucial role in generating thrust during the flutter kick, we speculated that delivering assistive torque to the hip via exoskeleton technology can effectively reduce muscle activation and gas consumption during the flutter kick in underwater diving \cite{Blake2013SpringPlus, Bhargava2004JoB, Walcott2012Bio}.

\begin{figure*}[htbp]
    \centering{
    \includegraphics[width=0.9\textwidth]{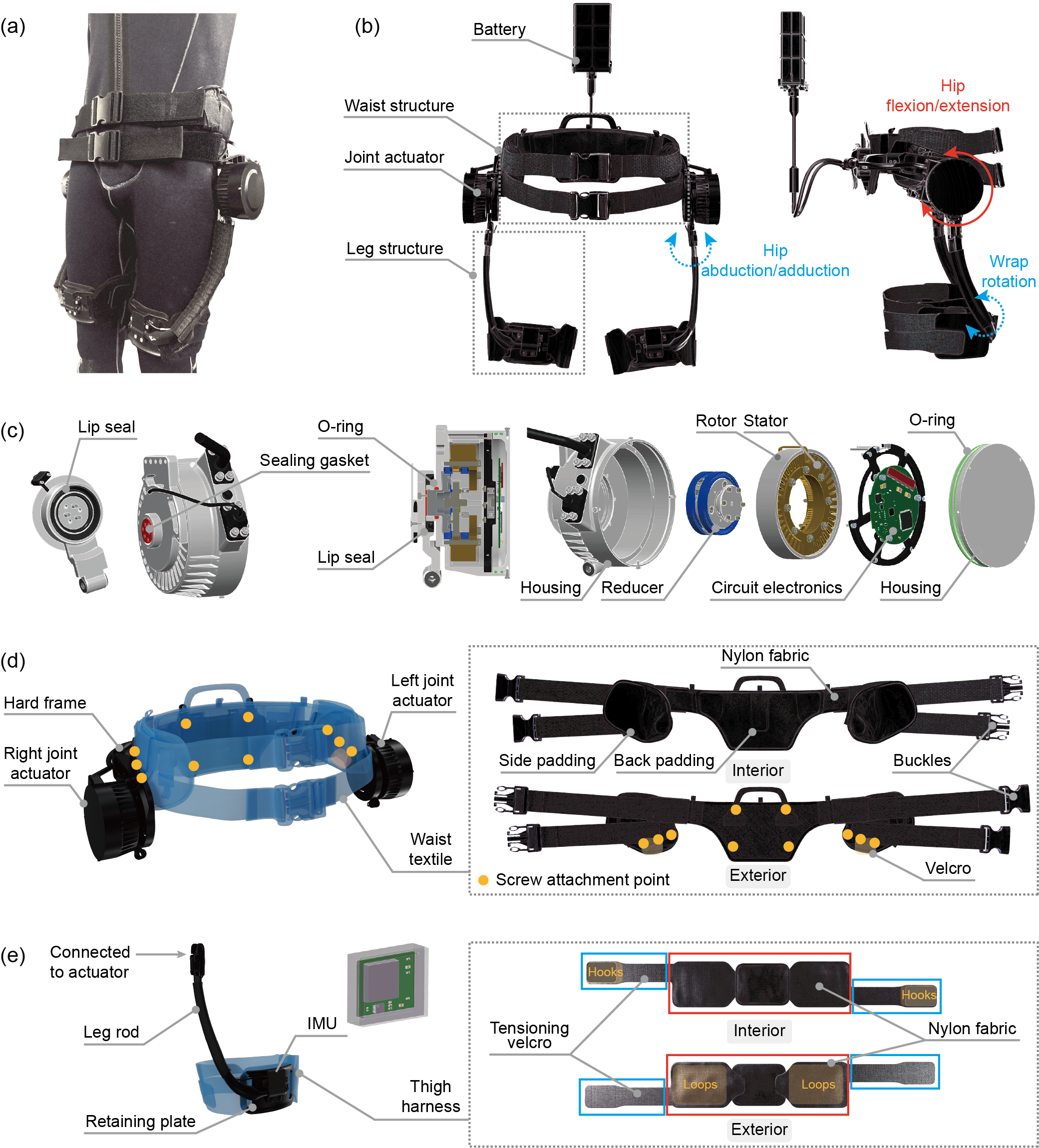}}
    \caption{DiveMate design. (a) A dummy wears the DiveMate. (b) DiveMate comprises two leg structures, two joint actuators, a waist structure, and a battery. It can be comfortably worn over the diving suit, and the user can freely perform hip abduction/adduction and flexion/extension. The actuation relies on the joint actuator, with the leg structure linked and transmitting torque to the user’s legs. (c) The joint actuator drives the leg structure to make a reciprocating motion. The inner structure of the joint actuator includes a stator, a rotor, a reducer, and circuit electronics. The sealing structure of the joint actuator contains a lip seal and O-rings. (d) The waist structure contains a hard frame and waist textile. Screw attachment points are placed on the frame and the textile to secure them together. (e) The leg structure contains a leg rod, a retaining plate, and a thigh harness. Inside the retaining plate, there is an IMU sensor. }
    \label{Fig_DiveMateDesign}
\end{figure*}  

\begin{table}[htbp]
\centering
\caption{Mass distribution of the DiveMate}
\fontsize{9}{10}\selectfont % 9pt字体，11pt行距
\label{tab:mass_distribution}
\begin{tabular}
{>{\raggedright\arraybackslash}m{0.18\textwidth} @{}
    >{\centering\arraybackslash}m{0.09\textwidth} @{}
    >{\centering\arraybackslash}m{0.09\textwidth} @{}
    >{\centering\arraybackslash}m{0.09\textwidth} @{}}
\toprule
Modular & Left side & Right side & Center \\
\midrule
Waist structure (kg) & / & / & 0.96 \\
Joint actuator (kg) & 1.12 & 1.12 & / \\
Thigh structure (kg) & 0.22 & 0.22 & / \\
Battery (kg) & / & / & 1.22 \\
\midrule
\textbf{Total weight (kg)} & \multicolumn{3}{c}{\textbf{4.86}}\\
\bottomrule
\end{tabular}
\end{table}

\subsection{Mechanical Design}
We developed a field-deployable diving exoskeleton, DiveMate, for real-world underwater application (Fig. \ref{Fig1}).  A schematic of the DiveMate components is provided in Fig. \ref{Fig_DiveMateDesign}. The proposed exoskeleton integrates seamlessly with standard SCUBA gear, enabling self-contained underwater exploration. The DiveMate was designed to assist divers’ hip extension and flexion during flutter kicking. The total weight is 4.86 kg, symmetrically located on the waist while wearing (Table \ref{tab:mass_distribution}). The DiveMate features a modular design comprising dual-sided joint actuators, a waist structure, dual-sided leg structures, and a battery (Fig. \ref{Fig_DiveMateDesign}\textcolor{blue}{a} and \ref{Fig_DiveMateDesign}\textcolor{blue}{b}). It can be easily donned over clothing around the waist and legs before putting on SCUBA equipment.

% \subsection{Actuator design}
The joint actuator was specifically designed for protection against water and sediment in complex aquatic environments. The custom joint actuator consists of a brushless stator and rotor, a single-stage 9:1 gear reducer, internal circuit electronics (driver and controller), and housing (Fig. \ref{Fig_DiveMateDesign}\textcolor{blue}{c}). 
It has a peak torque of 30 Nm, a rated continuous torque of 22 Nm, and a no-load speed of 12.5 rad/s. The actuators installed on the waist structure drive the leg to reciprocate through the output shaft. Using a lip seal and an O-ring between the shaft and housing, the sealing design ensures the actuators are watertight. Even when used in turbid waters, it can prevent mud and sand from entering the sealing ring and causing malfunctions.
 
% \subsection{Waist structure}
The waist structure contains an integral hard frame and a compliant waist textile, with the two components connected at multiple points (Fig. \ref{Fig_DiveMateDesign}\textcolor{blue}{d}). The width of the frame can be easily adjusted to accommodate different body shapes. The hard frame is connected to the human body via the waist textile, which maintains overall stability while reducing the pressure exerted by the exoskeleton on the body. The waist textile consists of three parts: the main belt, the padding, and the buckles. The main belt features an inner layer of hard plastic, which increases the lateral rigidity to counteract deformation caused by the exoskeleton torque. It also features a Velcro loop interface, offering easy attachment of padding. There are three paddings in total to fit the human body's curve, located on both sides and the back of the waist. The buckles are positioned on the front, making it convenient to adjust the belt's tightness.

% \subsection{Leg structure}
The leg structure contains a carbon fiber (T700) leg rod, a retaining plate, and a thigh harness (Fig. \ref{Fig_DiveMateDesign}\textcolor{blue}{e}). One end of the leg rod is connected to the joint actuator, and the other end is connected to the retaining plate through hinges, maintaining the freedom of hip joint abduction/adduction. Inside the retaining plate, there is a packaged Inertial Measurement Unit (IMU; Xsens MTi-3, Movella, USA) used to collect kinematic information of the user’s thigh. The retaining plate is fixed to the thigh through the tensioning thigh harness, which consists of a nylon fabric and two stretchable Velcro fasteners. The exterior of the fabric is the loop interface of the Velcro, and the interior of the elastic Velcro is the hook interface of the Velcro. Users can easily stretch the Velcro with both hands and stick it to the exterior surface of the fabric when wearing the DiveMate. The force generated by the joint actuator is transmitted to the harness via the leg rod, providing auxiliary torque that assists the wearer’s hip flexion and extension.

% \subsection{Power Component}
The DiveMate is powered by a portable lithium polymer battery (24V, 3000mAh). The battery housing is made of a corrosion-resistant aluminum alloy, and the end cover is sealed with an O-ring. The housing is installed using a special fixing clip with a rotating handle that can be easily attached to the SCUBA tank. The portable battery can allow about 1 hour of continuous diving on a single charge and can be easily replaced if it runs out during use. Battery endurance is estimated from measured battery consumption during a 10-min continuous underwater kicking test at the rated assistance condition.

\begin{figure*}[htbp]
    \centering
    \includegraphics[width=0.9\textwidth]{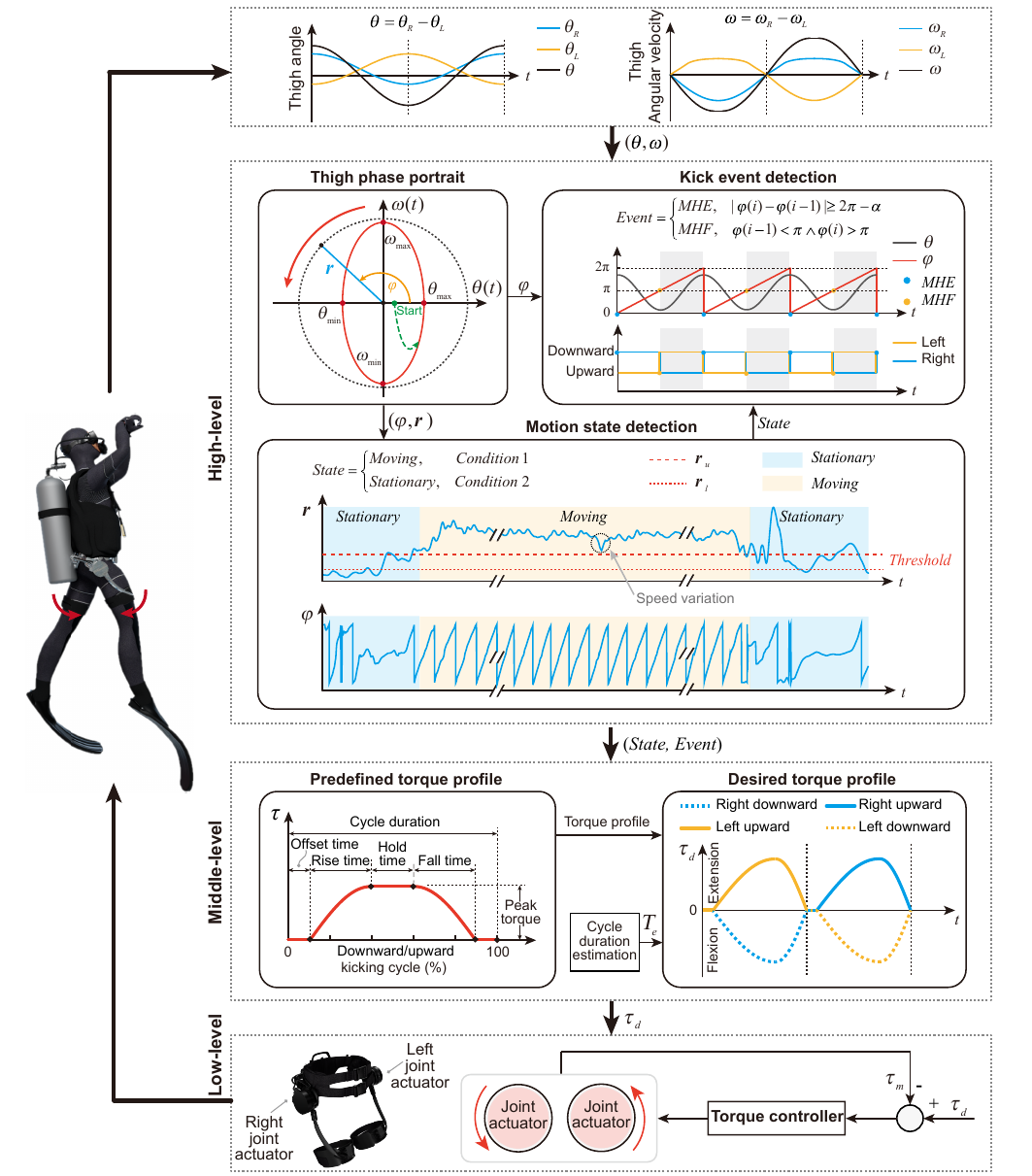}
    \caption{Control framework of DiveMate. It is designed to adapt to human aquatic locomotion. It targets hip joint assistance during alternating hip flexion-extension of underwater flutter kicking. The high-level controller implements detection of diving motion state and kick events to determine when assistance should be activated or deactivated. Polar angle ($\varphi$) and polar radius ($r$) of the thigh phase portrait $(\theta(t),\omega(t))$ are calculated and utilized to detect motion state (Stationary state and Moving state). If a human enters the Moving state, the kick-event detection algorithm determines when to apply assistance during each phase. The middle-level controller generates desired assistance torque profiles for DiveMate. We establish a predefined torque curve $\tau$ characterized by temporal parameters and a magnitude parameter. The desired torque profile is generated by proportionally scaling the predefined profile according to the duration of the next kicking cycle. This duration is estimated as a weighted average of the three preceding cycles. The low-level controller implements real-time torque regulation through a closed-loop architecture. }
    \label{Fig_Controller}
\end{figure*}
\section{Control System}
DiveMate needs to adapt to the variable pace of diving kicks encountered in real-world environments. Based on human diving kinematics and lower-limb muscle activation patterns, we designed a hierarchical control framework for DiveMate to adapt to human aquatic locomotion. It targets bidirectional assistance during alternating bilateral hip flexion-extension of underwater flutter kicking. Within the control architecture, signals flow unidirectionally from sensors through the control unit to actuators, with the resultant output torque ultimately applied to the human body, establishing a closed-loop operational cycle. The framework adopts a three-tiered architecture: the high-level control layer detects assistance conditions using real-time kinematic data from the bilateral limbs; the middle-level layer generates reference assistance torque profiles in a time-series adaptive manner; and the low-level control layer executes the precise delivery of assistance torque. The control framework is depicted in Fig. \ref{Fig_Controller}.

\subsection{High-Level Controller}
The high-level controller acquires bilateral lower-limb kinematic signals, including thigh angles and angular velocities, from IMUs mounted on the leg structure's retaining plate. Then, we developed a two-class detection algorithm consisting of diving motion state detection and kick event detection to determine when assistance should be activated or deactivated.

During underwater flutter kicking, human bilateral lower limbs exhibit periodic sinusoidal patterns in both angle $\theta(t)$ and angular velocity $\omega(t)$ of the hip joint, with $\omega(t)$ lagging $\theta(t)$ by a phase shift of $\pi/2$. These two variables form an ellipse rotating around the origin in the phase portrait $(\theta(t),\omega(t))$ coordinate \cite{Gregg2017TNSRE} (High-level of Fig. \ref{Fig_Controller}). 

The angle difference $(\theta(t)=\theta_R (t)-\theta_L (t)$ and angular velocity difference $(\omega(t)=\omega_R (t)-\omega_L (t))$ in the sagittal plane between the two legs are used to construct the thigh phase portrait. To reduce noise-induced destabilization, the Kalman filter is used to smooth $\theta(t)$ and $\omega(t)$. We reconstruct a monotonically increasing kicking phase portrait according to the filtered signal. 
The polar angle $\varphi(t)$ from the phase portrait is an indication of the progression of the kicking phase and is computed through the following equation
\begin{equation}
\varphi(t)=atan2(\hat{\omega}(t),\hat{\theta}(t))+\pi
\end{equation}  

where $\hat{\theta}(t)$ and $\hat{\omega}(t)$ are the hip angle difference and velocity difference after standardization; the addition of $\pi$ was done to ensure that the range of $\varphi(t)$ is between 0 and $2\pi$. 

Standardization through scaling and shifting of $\theta(t)$ and $\omega(t)$ is needed to enhance phase variable circularity \cite{Gregg2017CCTA}. Both variables are shifted about the origin of the hip phase portrait, and $\theta(t)$ is re-scaled to match the amplitude of $\omega(t)$:

\begin{equation}
\hat{\theta}(t)=z(t)\cdot(\theta(t)+\gamma(t))
\end{equation}

\begin{equation}
\hat{\omega}(t)=-(\omega(t)+\Gamma(t))
\end{equation}

where $z(t)$ is the scale parameter, and $\gamma(t)$ and $\Gamma(t)$ are the shift parameters calculated from the angle and velocity, respectively.

\begin{equation}
z(t) = \lvert\frac{\omega_{max}(t)-\omega_{min}(t)}{\theta_{max}(t)-\theta_{min}(t)}\rvert
\end{equation}

\begin{equation}
\gamma(t) = -\frac{\theta_{max}(t)+\theta_{min}(t)}{2}
\end{equation}

\begin{equation}
\Gamma(t) = -(\frac{\omega_{max}(t)+\omega_{min}(t)}{2})
\end{equation}

where the maximum and minimum values of $\theta(t)$ and $\omega(t)$ are related to the last kicking cycle.

Before a complete kicking cycle was available, the normalization extrema were initialized using preset values obtained from preliminary underwater tests ($\theta_{max}$=30, $\theta_{min}$=-30, $\omega_{max}$=150, $\omega_{min}$=-150). The extrema were updated only after valid MHF and MHE events were detected. If an event was missed, the most recently accepted extrema were retained. 

The polar radius $r(t)$ in phase portrait is defined as:

\begin{equation}
r(t) = \sqrt{\hat{\theta}(t)^2+\hat{\omega}(t)^2}
\end{equation}

This phase portrait-based approach originates from the general control theory of oscillatory dynamical systems \cite{Gregg2017TNSRE,Westervelt2018Fdbk}. In this study, we used the difference of bilateral thigh angle and the difference of bilateral kicking angular velocity to reconstruct the thigh phase portrait and to calculate the polar angle and polar radius in real-time. The difference is defined as the signal of the right leg subtracting that of the left leg. Using the difference can detect bilateral hip motion intentions and simultaneously apply bilateral assistance torques. It can help maintain kicking stability, given the buoyancy of the water.

\subsubsection{Motion State Detection}
We first implemented motion state detection, defining two motion states: $Stationary$ state and $Moving$ state. Quintero et al. \cite{Gregg2017CCTA} performed start and stop detection by comparing phase portrait coordinates with an elliptical boundary centered around the phase portrait origin \cite{Gregg2017CCTA}. Similarly, we can distinguish two motion states by polar radius thresholding. If the polar radius exceeds the defined threshold, the human is considered to enter the $Moving$ state; otherwise, as being in the $Stationary$ state.  Continuous, steady diving locomotion is a prerequisite for assistance activation. Although polar radius thresholding can classify motion states, abnormal fluctuations in polar radius due to low-regularity kicking can lead to unstable detection. 

To improve the accuracy and robustness of detection, this study integrated polar radius $(r)$ and polar angle $(\varphi)$ in the detection algorithm. The $\varphi$ increases linearly from 0 to $2\pi$ in each normal kicking cycle but exhibits irregular variation during the transition between two states. To differentiate between linear progression and irregular variations for polar angle, we perform linear regression on $\varphi$ within a sliding temporal window. The goodness of fit for the linear regression is evaluated using the coefficient of determination $(R^2)$: 
\begin{equation}
R^2_\varphi  = 1-\frac{SSE}{SST}
\end{equation}

where $SSE$ is the sum of squared errors between predicted and actual polar angle values, and $SST$ is the total sum of squares between actual values and the mean of the polar angle in the sliding temporal window. The sliding window is reset whenever the polar angle wraps from $2\pi$ to 0. When $\varphi$ increases linearly, $R^2_\varphi$ approaches 1; otherwise, irregular fluctuations result in a lower $R^2_\varphi$. Additionally, the regression coefficient $k_\varphi$, which indicates the kicking frequency, necessarily satisfies the physiological movement range during steady swimming. 

Consequently, to determine whether a human enters relatively steady swimming, the $R^2_\varphi$ and $k_\varphi$ for the linear regression of $\varphi$ are utilized to assist motion state detection. Briefly, if the $r$ exceeds the predefined radius threshold $r_u$, $R^2_\varphi$ exceeds the predefined threshold $R_0$, and the slope of linear regression $k_\varphi$ belongs to $(k_1,k_2)$, motion state is thought to transform from $Stationary$ state to $Moving$ state; if the $r$ is less than the predefined radius threshold $r_l$ or $R^2_\varphi$ is less than $R_0$ or the $(k_\varphi)$ does not belong to $(k_1,k_2)$, motion state is thought to transform from $Moving$ state to $Stationary$ state; otherwise, the current motion state is maintained. 

\begin{equation}
state=
\begin{cases}
Moving, & \mathrm{if} \  r\in[r_u,\infty) \ \mathrm{and} \ R^2_\varphi\in(R_0,1]\\
\  &\mathrm{and} \ k_\varphi\in(k_1,k_2)
\\[0.5ex]
Stationary, & \mathrm{if} \  r\in[0,r_l) \ or \ R^2_\varphi\in(0,R_0] \\
\  & or \  k_\varphi\notin(k_1,k_2 )
\\[0.5ex]
\end{cases}
\end{equation}

Experimental validation results are provided in Supplementary Fig. \textcolor{blue}{S1}, including locomotion transitions between $Stationary$ and $Moving$ states and moving speed variation. The implementation parameter for the sliding window length was set to 1000 ms. The dual threshold parameters $r_u$ and $r_l$ for polar radius are empirically set to 100 and 50, respectively. $R_0$ is configured at 0.8. $k_1$ and $k_2$ are configured at $4\pi/5$ and $2\pi$, respectively, since the normal cycle duration is from 1.0$s$ to 2.5$s$. Further details regarding the physical meaning and selection rationale of these parameters are provided in Supplementary Text \textcolor{blue}{S1}.

\subsubsection{Kick Event Detection}
If the human enters the $Moving$ state, the kick event detection algorithm determines when to apply assistance during each kicking phase. The polar angle $\varphi(t)$ is used as the input of the algorithm. We defined two critical kick events and thereby a kicking cycle is divided into two conjunctive phases (downward kicking and upward kicking): The first event ($\varphi=0$ or $\varphi=2\pi$) refers to the maximum hip extension timing of the right leg (MHE event), which is defined as the timing of the upward-to-downward transition of the right leg, while the left leg performs the opposite movement. The second event ($\varphi=\pi$) refers to the maximum hip flexion timing of the right leg (MHF event), which is defined as the timing of the downward-to-upward transition of the right leg, while the left leg performs the opposite movement. 

During the actual deployment of the algorithm, the sampling rate and acquisition error of the sensor can only make $\varphi$ as close as possible to 0 and $\pi$. Consequently, an MHE event occurs when the absolute value of the previous $\varphi$ minus the current $\varphi$ is larger than $2\pi-\alpha$, where $\alpha$ is $\pi/18$; an MHF event occurs when the current $\varphi$ is larger than $\pi$, and the previous $\varphi$ is less than $\pi$. In this study, seven healthy participants validated the event detection algorithm, and the results show that the overall detection accuracy was 99.6 ± 0.2\% and the overall detection delay was 53.2 ± 2.1 ms across all the subjects (Supplementary Fig. \textcolor{blue}{S2} and Supplementary Table \textcolor{blue}{S1}).

\subsection{Middle-Level Controller}
The middle-level controller generates desired assistance torque profiles for DiveMate. The desired profile is updated at each kick event. We aim to provide synchronized bilateral hip assistance with equal torque amplitude to avoid upper-body imbalance caused by asymmetric bilateral assistance torques. Therefore, only one directional torque profile needs to be designed, with its sign reversed for the contralateral leg and opposite kicking phase. Here, we establish a predefined torque curve $\tau$, and the parameters of $\tau$ consist of temporal parameters (offset time, rise time, hold time, fall time) and a magnitude parameter (peak torque). Temporal parameters are expressed as percentage ratios relative to the kicking cycle. The designed motion profiles for the rise and fall parts employ trigonometric functions with a period of $\pi/2$. 
The desired torque profile is generated by proportionally scaling the predefined torque profile according to the duration of the next kicking cycle. This duration $T_e$ is estimated as a weighted average of the three preceding cycles, applying temporal smoothing to enhance prediction stability. It is calculated by
\begin{equation}
    T_e= \alpha_1 T_1+\alpha_2 T_2+\alpha_3 T_3
\end{equation}
where $T_1$, $T_2$, $T_3$ are the last three preceding cycle durations, and $\alpha_1$, $\alpha_2$, $\alpha_3$ are the corresponding weight coefficients for the respective cycle times. Half of the estimated cycle duration determines the current assistance cycle (upward/downward kicking). The generated desired assistance torque profiles for the current kicking are subsequently transmitted to the next-level controller.
Implementation parameters for the flexion and extension phases are identical, and the temporal parameters are set to [10, 60, 0, 30]. 
The predefined torque-profile shape was informed by our previous biomechanical analysis of flutter kicking \cite{yang2026JOB}, and its parameterization is described in Supplementary Text \textcolor{blue}{S2}.

The peak assistance torque magnitude was manually selected during the familiarization procedure with DiveMate. During familiarization, participants briefly experienced seven peak assistance torque settings (6, 10, 14, 18, 22, 26, and 30 Nm). The 26- and 30-Nm settings exceeded the actuator’s rated continuous torque and were therefore tested only briefly under supervision and were not available for the formal endurance trials. In this study, all participants selected a peak assistance torque of 22 Nm during the formal experiments. 

The weighted average method balances recency and historical continuity in kinematic prediction, with weighting coefficients [$\alpha_1$, $\alpha_2$, $\alpha_3$] = [0.500, 0.333, 0.167]. The mean absolute error of cycle duration is about 0.09 s (5.67\% of cycle duration) (see Supplementary Table \textcolor{blue}{S2}).

\subsection{Low-Level Controller}
The low-level controller implements real-time torque control through a proportional-integral-derivative (PID) closed-loop architecture. A feedback torque loop compares the actual torque of the motor $\tau_m(t)$ (motor-current estimation) with the desired torque $\tau_d(t)$ extracted from the previous layer. This operational sequence begins with high-frequency torque acquisition, followed by error calculation via digital signal subtraction, then generates corrective voltage commands via a PID controller, and ultimately implements current regulation by applying pulse-width-modulated (PWM) voltage to motor drivers. The controller dynamically adjusts the motor current output to achieve target torque tracking based on the error between the measured and desired torque. This continuous process operates at 20 kHz, maintaining torque fidelity within ±2.5 \% of the reference values during dynamic load variations. 

Real-time data of DiveMate during diving is shown in Supplementary Video \textcolor{blue}{3}. Kinematic signals (thigh angles), kick events, kicking phases (downward kicking and upward kicking), and scaled assistive torque outputs are real-time mapped onto corresponding leg kicking during underwater locomotion.
\section{Experiments}

\subsection{Subjects Information}
 Seven recruited participants (detailed information in Table \ref{tab:participants_attributes}) performed underwater diving experiments with standard SCUBA diving equipment in an indoor swimming pool (Supplementary Fig. \textcolor{blue}{S3}). All participants were first-time users of underwater exoskeletons, with 4 having experience with land-based exoskeletons. No participants reported any injuries that could affect their diving. All participants provided written informed consent and written consent to publish identifiable photos and videos. The Ethical Review Board of the Institute of Automation, Chinese Academy of Sciences approved the experimental protocol.

\begin{table}
\centering
\caption{Attributes of participants}
\label{tab:participants_attributes}
\fontsize{8}{10}\selectfont % 9pt字体，11pt行距
\scalebox{1.0}{
\begin{tabular}{  
    >{\centering\arraybackslash}m{0.075\textwidth} @{}
    >{\centering\arraybackslash}m{0.065\textwidth}   @{}
    >{\centering\arraybackslash}m{0.065\textwidth} @{}
    >{\centering\arraybackslash}m{0.065\textwidth}  @{}
    >{\centering\arraybackslash}m{0.065\textwidth}   @{}
    >{\centering\arraybackslash}m{0.122\textwidth}  @{}
} 
\toprule
\multirow{2}{*}{Participant} & Age & \multirow{2}{*}{Gender} & Height & Weight  & Experience  \\
                             & (years) &   &  (cm)  &   (kg)  & with exoskeletons\\ 
\midrule
Subject 1 & 34 & Male & 160 & 70 & Yes \\
Subject 2 & 35 & Male & 168 & 73 & Yes \\
Subject 3 & 23 & Male & 181 & 76 & No \\
Subject 4 & 22 & Male & 179 & 98 & No \\
Subject 5 & 25 & Male & 188 & 105 & No \\
Subject 6 & 31 & Male & 162 & 87 & Yes \\
Subject 7 & 25 & Male & 182 & 93 & Yes \\
\midrule
Mean & \(27.9 \) & \multirow{2}{*}{7M} & \(174.3 \) & \(86.0 \) & \multirow{2}{*}{4Y, 3N} \\
 \(\pm\) SD & \(\pm\) 5.4 &  & \(\pm\) 10.9 & \(\pm\) 13.4 &  \\
\bottomrule
\end{tabular}
}
\end{table}

\subsection{Experiment Platform}
\subsubsection{Swimming Pool Facility}
The underwater diving test was conducted in an indoor swimming pool (length: 25 m, width: 13.3 m, depth: 1.6 m, water temperature: 26 ± 0.5℃) (Supplementary Fig. \textcolor{blue}{S3}). To normalize participants’ swimming trajectory and avoid the effects of right-angle turns, we set an octagonal swimming path (56.06 m per cycle) using lane lines, the drainage channel, and four weight belts on the bottom of the pool. The entry point is the beginning of the lane line near the ladder (blue hollow point in Supplementary Fig. \textcolor{blue}{S3}).

%浮漂设计
\subsubsection{Communication Buoy}
A communication buoy is designed to serve as a relay from an underwater wearable device to a telemetry laptop at the poolside in this study (Fig. \ref{Fig_Results_Endurance}\textcolor{blue}{a} and \textcolor{blue}{c}). It is used to transmit commands from the laptop and monitor the test process conveniently. The communication buoy consists of a circuit board, a waterproof communication interface, and a buoyancy housing. A 32-bit microcontroller unit (MCU; STM32H743, STMicroelectronics, Switzerland) communicates with the exoskeleton via a universal asynchronous receiver/transmitter (UART) to a recommended standard 422 (RS-422) converter. The MCU communicates with the modified pressure gauge via a UART-to-RS-485 converter and synchronizes data from those devices. Simultaneously, the MCU communicates with a telemetry laptop via a Wi-Fi module (USR-C322, USR IOT) at 100 Hz for real-time synchronous data visualization and recording.

%气压表设计
\subsubsection{Modified Pressure Gauge for Gas Consumption Measurement}
To enable continuous monitoring of gas consumption during diving, we modified a commercial pressure gauge (CRESSI DIGI, CRESSI, Italy) to transmit real-time gas pressure signals (Supplementary Fig. \textcolor{blue}{S4}). We designed a Bluetooth transmitter with a Bluetooth chip (nrf52832, Nordic, Norway), which is integrated into the original pressure gauge to read the pressure signal via the inter-integrated circuit (IIC) from the pressure transmitter. A Bluetooth receiver wirelessly acquires data from the modified pressure gauge. The data is transmitted from the Bluetooth receiver to the underwater communication buoy through the RS-485 protocol. Although the Bluetooth receiver and transmitter require proximity underwater ($<$15 cm), this approach keeps the original housing of the pressure gauge, avoiding structural modifications that might destroy waterproofing. 
To quantify gas consumption levels, gas pressure-time data $P$ (unit: bar) from the start to the stop of swimming are recorded. According to the ideal gas law, the relationship between the pressure change $\Delta P$ and gas consumption $V$ is
\begin{equation}
V \cdot P_s=\Delta P \cdot V_{Tank}
\end{equation}
where $V_{Tank}$ represents the volume of the self-contained pressure tank (2 L), and $P_s$ represents the standard atmospheric pressure (1.01325 bar). Gas consumption V is
\begin{equation}
V=\frac{\Delta P}{P_s} \cdot V_{Tank}
\end{equation}
The negligible change in the swimming pool, coupled with the tank being permanently submerged, meant that temperature variations were omitted in the gas consumption calculations using the gas law. The gas consumption rate is defined as the absolute value of the slope coefficient obtained by least-squares fitting (Supplementary Fig. \textcolor{blue}{S4}).

\subsection{Experiment Protocol}
A diver's endurance is characterized by travel distance and dive duration achievable with a fixed volume of self-contained breathing gas. This study designs a comparative experiment to evaluate improvements in endurance performance under exoskeleton assistance across two conditions. Participants perform flutter kicks continuously while wearing the DiveMate, with assistance turned on (W-Assist), and with assistance turned off and thigh harnesses unfastened (Wo-Assist). The reason for not removing the exoskeleton in the Wo-Assist condition is to maintain the same weight, body shape, and buoyancy across both conditions. If the overall weight changes, the diver must establish neutral buoyancy again by inflating or deflating the BCD. In addition, unfastening the thigh harnesses frees the leg swing from the resistance torque imposed by the actuator unit's mechanical friction and inertia. The full experimental procedure, including the measurement of resting gas consumption, is shown in Supplementary Video \textcolor{blue}{4}.

%wgg熟悉过程
To minimize variability arising from prior device experience, all participants were recruited as first-time users of the underwater exoskeleton. However, to mitigate any potential disparities associated with the familiarization process, a dedicated familiarization process was implemented prior to data collection. During this process, participants were allocated sufficient time to adapt to the device. Specifically, each participant was required to wear the exoskeleton and perform continuous flutter kicks along the yellow straight path in an indoor swimming pool (Supplementary Fig. \textcolor{blue}{S3}). Participants were encouraged to repeat exoskeleton-assisted kicking in the experimental setup until they gained confidence in its operation. The familiarization process was conducted the day before the formal experiment to ensure participants had adequate rest before starting the experimental tasks, thereby mitigating fatigue associated with exoskeleton familiarization. 

%穿戴设备详细介绍
Before each underwater experiment, all participants donned standard diving equipment (Fig. \ref{Fig_Results_Endurance}\textcolor{blue}{b}), which consists of a buoyancy control device, two diving regulators with first and second stages, a diving mask, diving shoes (SCUBAPRO, America), diving fins (MARES, Italy), pressure gauges (CRESSI, Italy), and two SCUBA diving tanks (2 liters). One tank was used to adjust neutral buoyancy before movement, while the other supplied gas for participants’ breathing during the formal test (all gas consumption measurements and analyses reported in this paper are from this gas tank). In addition, wetsuits in different sizes were provided, and participants could choose whether to wear one for the best fit. Diving with DiveMate assistance is shown from the top and back views in Supplementary Video \textcolor{blue}{5}.

After putting on all the equipment and the DiveMate, the subjects were instructed to jump into the water and establish neutral buoyancy by adjusting the BCD, ensuring their bodies did not touch the pool bottom or rise above the water. Then, we measured resting gas consumption while participants remained submerged, without additional movement, for 5 minutes (±30 seconds). Next, participants completed a 5-minute warm-up under both W-Assist and Wo-Assist conditions. Before departure, the diver used tank 1 to breathe and adjust neutral buoyancy. After the adjustment, the diver changed to the breathing valve of tank 2 and departed. Afterward, participants were required to perform continuous flutter kicks from the start point along the predefined path at a self-comfortable speed under W-Assist and Wo-Assist conditions. Each condition lasted until the tank-pressure decrease reached approximately 150 bar (±10\%) (Fig. \ref{Fig_Results_Endurance}\textcolor{blue}{b}). Participants were instructed to stop at the nearest vertex of the predefined octagonal path to easily record the travel distance. The order of the two conditions was randomized, and a minimum 30-minute rest interval on land separated them to mitigate fatigue-related effects. 

A researcher in the water provided instructions for the start and end of each underwater test. More importantly, this researcher provided safety supervision and could initiate rescue procedures in an emergency. Another researcher at the poolside sends control commands to the exoskeleton remotely via a laptop and a custom Wi-Fi communication buoy (Fig. \ref{Fig_Results_Endurance}\textcolor{blue}{a}). Meanwhile, gas consumption and exoskeleton data are monitored and recorded in real-time. Muscle activity and kinematic data from the wireless EMG sensor modules were recorded locally for offline analysis. 

\subsection{Data Analysis}
\subsubsection{Gas Consumption and Endurance} 
Gas consumption is obtained from the tank pressure recordings. Tank pressure–time data were first fitted using least squares to obtain the tank-pressure decrease rate. The gas consumption rate at standard atmospheric pressure was then calculated (Supplementary Fig. \textcolor{blue}{S4}). Then, we can obtain the resting gas consumption rate and gross gas consumption rates in the resting experiment and diving experiments under both W-Assist and Wo-Assist conditions, respectively. Moreover, the net gas consumption rate is defined as the gross rate minus the resting rate, thereby approximating the gas consumption attributable to locomotion by removing baseline consumption.

To evaluate the effectiveness of DiveMate in improving divers’ endurance, the principal indicators are travel distance and dive duration while consuming a given volume of breathing gas (tank-pressure decrease of 150 bar). The gas consumption of transport (CoT) is obtained by dividing the gas consumption by the recorded travel distance. Since endpoint gas consumption varied by approximately ±10\% across trials, the directly recorded travel distance and dive duration (raw values) were affected by this inconsistency. We therefore normalized both metrics to a common tank-pressure decrease of 150 bar. Travel distance was calculated by dividing the given gas volume by CoT, and dive duration was calculated by dividing the given gas volume by the gross gas consumption rate. Unless otherwise specified, “travel distance” and “dive duration” hereafter refer to these normalized values, whereas the directly recorded trial-level values are termed ``raw travel distance'' and ``raw dive duration''. The directly measured variables and the metrics derived from them are summarized in Supplementary Fig. \textcolor{blue}{S5}, which also illustrates the computational relationships among the reported outcome measures.

\subsubsection{Kinematics and Muscle Activity} 
To evaluate muscle activity under W-Assist and Wo-Assist conditions, surface EMG signals were recorded from the rectus femoris, biceps femoris, and semitendinosus muscles at 2 kHz using a wireless system (MiniWave, Cometa, Italy). The skin surface of the participants was shaved and cleaned with alcohol prior to EMG electrode fixation. Electrodes were placed in accordance with SENIAM guidelines and sealed with a transparent, waterproof film dressing (Tegaderm film, 3M Company, USA). Kinematic signals were from DiveMate (recorded at 100 Hz) and the IMU integrated into the EMG sensors. The surface EMG signals and kinematic data were first linearly interpolated to their respective timestamps, then synchronized by aligning the measured accelerations. Raw EMG signals were high-pass filtered (using a zero-lag 2nd-order Butterworth filter with a cutoff frequency of 20 Hz), rectified, and low-pass filtered (using a zero-lag 2nd-order Butterworth filter with a cutoff frequency of 6 Hz) to create muscle-activity linear envelopes. The resulting EMG linear envelope was then normalized by the maximum activation recorded in the Wo-Assist condition, and the root mean square (RMS) of muscle activity and peak muscle activity were calculated. We also counted the temporal occurrence of peak muscle activities.

In this study, a kicking cycle was considered to be the period between subsequent maximal hip extensions of the diver’s right leg. The maximum hip extension (MHE) points were detected from the thigh angle profile of the diver’s right leg. Then, we employed these MHE points to divide the kicking cycles. The cycle duration denotes the period of a kicking cycle, and we also calculated the kicking frequency using the reciprocal of the cycle duration. All chronological data, including kinematics and muscle activities, were normalized to the kicking cycle (0-100\%). The duty factor of the downward kicking phase was calculated by dividing the downward kicking duration by the cycle duration. Furthermore, the diving speed was calculated by dividing the recorded travel length by the travel time.

\subsection{Data Statistics}
For each participant and experimental condition, a single participant-level value was calculated for each outcome metric by averaging across valid kicking cycles or analysis intervals, where applicable. Participant-level values were used as the units of statistical analysis. Therefore, all between-condition comparisons included seven paired observations (n = 7). Unless otherwise specified, group-level results are reported as the mean $\pm$ SEM.

Statistical analyses were performed using MATLAB R2023b (MathWorks, USA). For each metric, the normality of the paired within-participant differences between the W-Assist and Wo-Assist conditions was assessed using the Shapiro–Wilk test. When the paired differences did not significantly deviate from normality, the two conditions were compared using a two-sided paired t-test. When the normality assumption was not met, a two-sided Wilcoxon signed-rank test was used. Statistical significance was defined as $\alpha$ = 0.05. The statistical test used and the specific p-value for each comparison are reported in the corresponding Supplementary Tables.

\subsection{User Surveys} 
Participants completed a series of surveys to evaluate the sense of agency, comfort, and ease to use of the DiveMate after completion of all the experiments. The sense of agency survey (Table \ref{tab:agency_survey}) was adapted from the Sense of Agency Scale, which is a measure of consciously perceived control over one's mind, body, and the immediate environment \cite{ZhangXH2024NMI, Tapal2017FroPsy}. Participants were asked to complete a 10-item questionnaire, consisting of 6 positively framed and 4 negatively framed items, to assess their sense of control while using DiveMate. Each participant rated their level of agreement with the items on a scale ranging from 1 (strongly disagree) to 7 (strongly agree). The negatively framed items were recoded so that higher values indicate a higher sense of agency. In addition, participants were asked to complete a Comfort survey and a System Usability Scale survey (Table \ref{tab:comfort_usability_combined}) to assess comfort and ease of operation of the DiveMate. The comfort was adapted from the Orthotics and Prosthetics Users’ Survey \cite{Collins2022Nature,Heinemann2003ProsthetOI}, which acts as a self-report instrument for evaluating clinically useful outcomes of prosthetics and orthotics services. The usability measure was adapted from the System Usability Scale \cite{Collins2022Nature,Jordan1996CRCpress}, which uses a Likert scale to evaluate DiveMate's usability. Each participant rated their level of agreement with the items on a scale ranging from 0 (strongly disagree) to 4 (strongly agree).
\section{Results}

\begin{figure*}[htbp]
    \centering{
    \includegraphics[width=0.95\textwidth]{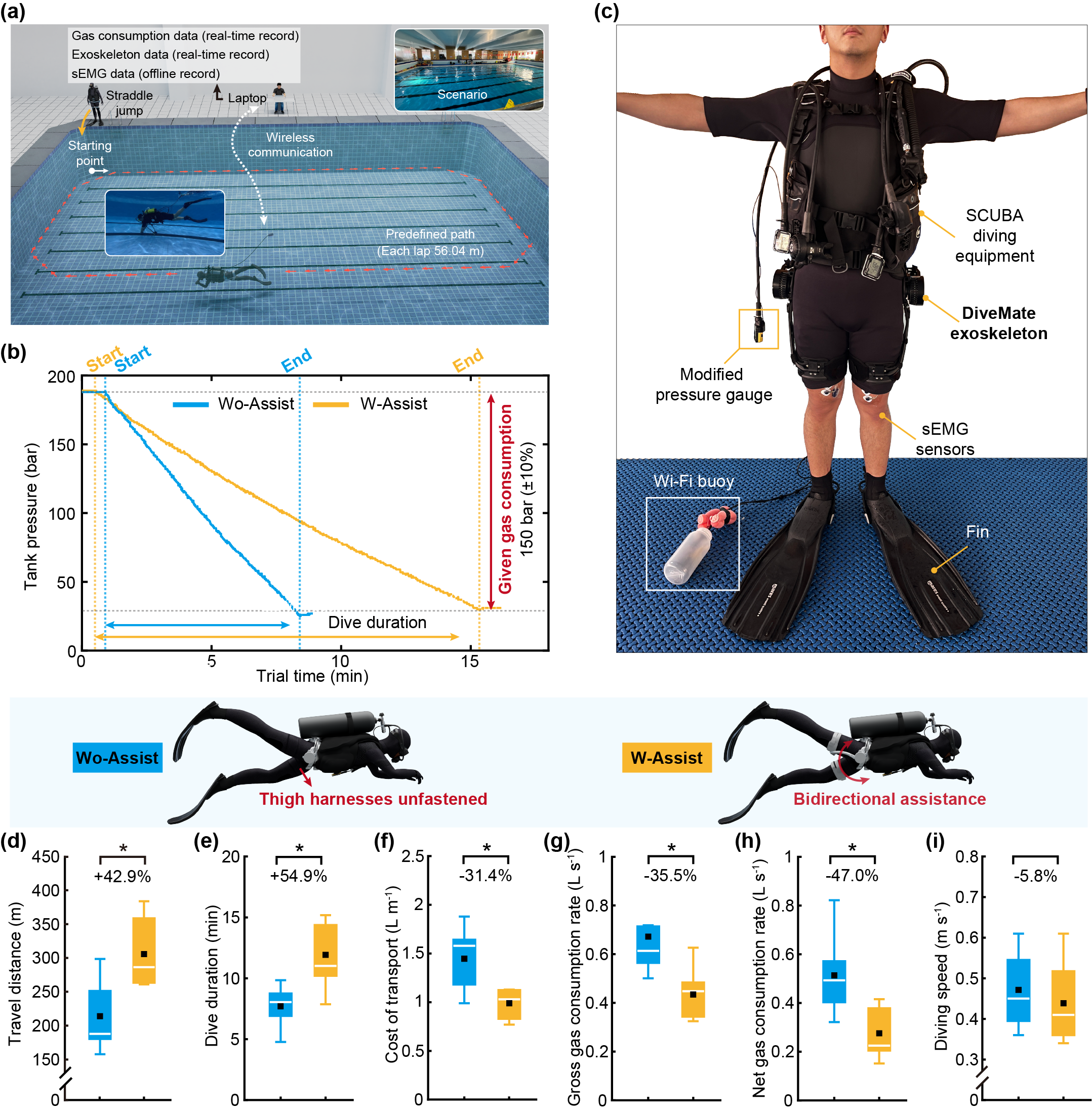}}
    \caption{Divers’ endurance assessment during diving. (a) Participants swam at their self-comfortable speed along the predefined path until the tank-pressure decrease reached approximately 150 bar (±10\%) under two conditions: wearing the DiveMate with assistance turned on (W-Assist) and wearing the DiveMate without assistance and thigh harnesses unfastened (Wo-Assist). Gas consumption data and exoskeleton data were wirelessly streamed via a custom Wi-Fi buoy system to a laptop for real-time monitoring and logging. Surface EMG signals were locally recorded within sensor modules for offline analysis. (b) Tank pressure-time data of one representative participant spanning the entire underwater swimming duration under both experimental conditions. (c) A subject donning the proposed diving exoskeleton and SCUBA diving equipment. (d and e) Travel distance and dive duration normalized to a given tank-pressure decrease of 150 bar were significantly increased with DiveMate assistance. (f) The gas consumption of transport was significantly reduced under the W-Assist condition compared to the Wo-Assist condition. (g and h) Both the gross gas consumption rate and the net gas consumption rate were significantly reduced with DiveMate assistance. (i) Diving speed did not differ significantly between the two conditions. In the box plot of (d to i), the center line represents the median, the black dot represents the mean, the box limits delineate the 25th and 75th percentiles, and the whiskers extend between the minimum and maximum of the data values (n = 7 for two conditions). Statistical significance and p-values are determined by a two-sided paired test; *p $<$ 0.05. }
    \label{Fig_Results_Endurance}
\end{figure*}

\subsection{Diving endurance}
DiveMate significantly improved diving endurance. When participants consumed a given amount of breathing gas (2 L gas tank with tank-pressure decrease of 150 bar), travel distance was significantly increased by 42.9\% with DiveMate (n = 7, p $<$ 0.05) (Fig. \ref{Fig_Results_Endurance}\textcolor{blue}{d}), and dive duration was significantly extended by 54.9\% (n = 7, p $<$ 0.05) (Fig. \ref{Fig_Results_Endurance}\textcolor{blue}{e}). Meanwhile, the CoT was significantly reduced by 31.4\% under the W-Assist condition compared to the Wo-Assist condition (n = 7, p $<$ 0.05) (Fig. \ref{Fig_Results_Endurance}\textcolor{blue}{f}). The human gross gas consumption rate was significantly reduced by 35.5\% (n = 7, p $<$ 0.05), and the net gas consumption rate was also significantly reduced by 47.0\% (n = 7, p $<$ 0.05) while further considering the resting gas consumption rate (Fig. \ref{Fig_Results_Endurance}\textcolor{blue}{g} and \textcolor{blue}{h}). Nonetheless, there is no significant difference in diving speed under the two conditions (n = 7, p = 0.265; Fig. \ref{Fig_Results_Endurance}\textcolor{blue}{i}). The above results across participants are detailed in Supplementary Tables \textcolor{blue}{S3}, \textcolor{blue}{S4}, and \textcolor{blue}{S5}. Raw recorded data are also provided in Supplementary Table \textcolor{blue}{S6}.

\subsection{Muscle activities}
\begin{figure*}[htbp]
    \centering{
    \includegraphics[width=0.95\textwidth]{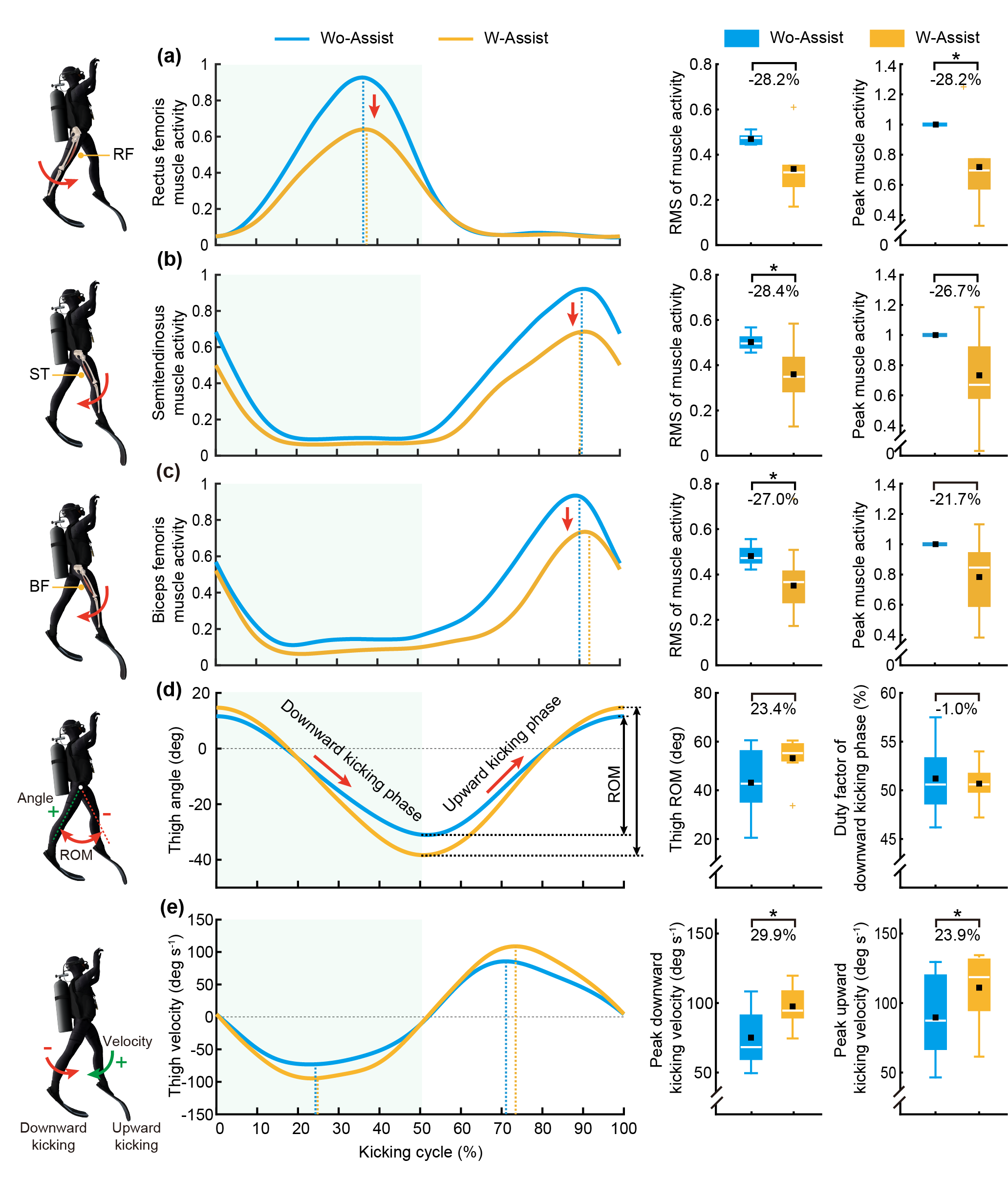}}
    \caption{Muscle activities and kinematics analysis under Wo-Assist/W-Assist conditions. (a to c) DiveMate reduced activities of the hip-related muscle groups (rectus femoris (a), semitendinosus (b), and biceps femoris (c)) without altering muscle activation patterns. The vertical dashed lines represent the temporal occurrence of peak muscle activation. (d) DiveMate does not restrict thigh motion, as indicated by the thigh range of motion (ROM). (e) Powerful kicking was found with increased thigh peak upward/downward velocity. In all time-normalized plots (a to e), the abscissa represents the percentage of the kicking cycle, with 0\% corresponding to maximum hip extension (MHE) of the right leg. In the box plots, the center line represents the median, the black dot represents the mean, the box limits delineate the 25th and 75th percentiles, and the whiskers extend between the minimum and maximum of the data values (n = 7 for two conditions). Statistical significance and p-values are determined by a two-sided paired test; *p $<$ 0.05.}
    \label{Fig_Results_AngleEMG}
\end{figure*}

DiveMate reduced activities of the hip-related muscle groups without altering muscle activation patterns. Normalized activation profiles of the hip flexor and extensor during flutter kicking are shown in the left column of Fig. \ref{Fig_Results_AngleEMG}\textcolor{blue}{a} to \ref{Fig_Results_AngleEMG}\textcolor{blue}{c}. The W-Assist condition led to reductions in the root mean square (RMS) of muscle activities, including the hip flexor (-28.2\% for the RMS of rectus femoris, n = 7, p = 0.0504) and hip extensor (-28.4\% for the RMS of semitendinosus, n = 7, p $<$ 0.05; -27.0\% for the RMS of biceps femoris, n = 7, p $<$ 0.05) (middle column of Fig. \ref{Fig_Results_AngleEMG}\textcolor{blue}{a} to \ref{Fig_Results_AngleEMG}\textcolor{blue}{c}). 

Under the W-Assist condition, the peak muscle activity of rectus femoris was significantly reduced by 28.2\% (n = 7, p $<$ 0.05) (right column of Fig. \ref{Fig_Results_AngleEMG}\textcolor{blue}{a}). There was no significant reduction in peak activity of the hip extensor (semitendinosus, n = 7, p = 0.058; biceps femoris, n = 7, p = 0.067), although there was a trend toward an average decrease of 26.7\% and 21.7\%, respectively (right column of Fig. \ref{Fig_Results_AngleEMG}\textcolor{blue}{b} and \ref{Fig_Results_AngleEMG}\textcolor{blue}{c}). In particular, six out of seven participants showed peak activity reductions in both the biceps femoris and semitendinosus under the W-Assist condition. 

Temporal occurrences of peak muscle activities (percentage of the kicking cycle) showed no significant differences between the two conditions (Vertical dashed lines in the left column of Fig. \ref{Fig_Results_AngleEMG}\textcolor{blue}{a} to \ref{Fig_Results_AngleEMG}\textcolor{blue}{c}). The above results across participants are detailed in Supplementary Table \textcolor{blue}{S7}.

\subsection{Diving kinematics}
More powerful kicking was found, and naturalistic hip kinematics were not restricted when using DiveMate. There is no significant difference in the thigh range of motion (ROM) between the two conditions (n = 7, p = 0.078) (Fig. \ref{Fig_Results_AngleEMG}\textcolor{blue}{d}), although the thigh ROM exhibited an increased trend (23.4\%). In particular, five out of seven participants increased their ROM under the W-Assist condition, and the remaining two participants maintained comparable ROM across experimental conditions. The duty factor of the downward kicking phase showed no significant difference between the two conditions (a 1.0\% decrease; n = 7, p = 1.000) (Fig. \ref{Fig_Results_AngleEMG}\textcolor{blue}{d}). However, the peak downward and peak upward kicking velocity were significantly higher with exoskeleton assistance than the Wo-Assist condition (29.9\% for peak downward kicking velocity, n = 7, p $<$ 0.05; 23.9\% for peak upward kicking velocity, n = 7, p $<$ 0.05) (Fig. \ref{Fig_Results_AngleEMG}\textcolor{blue}{e}), and the temporal occurrences of the peak downward and upward kicking velocity (percentage of the kicking cycle) have no significant differences under two conditions (the peak downward kicking velocity, n = 7, p = 0.688; the peak upward kicking velocity, n = 7, p = 0.294) (Vertical dashed lines in the left column of Fig. \ref{Fig_Results_AngleEMG}\textcolor{blue}{e}). The above results across participants are detailed in Supplementary Table \textcolor{blue}{S8}.

%%%%%%%%%%%%%%%%%%%%%%%%%%%%%%%%%%%8pt
\begin{table}[htbp]
\centering
\caption{Survey results on the sense of agency for exoskeleton participants}
\label{tab:agency_survey}
\fontsize{8}{10}\selectfont % 9pt字体，11pt行距
\begin{tabular}{  
>{\raggedright\arraybackslash}m{0.395\textwidth} @{}
    >{\centering\arraybackslash}m{0.075\textwidth}   @{}}
\toprule
\textbf{Question text (†: This item is inversely coded; 7 = Strongly Agree, 6 = Agree, 5 = Rather Agree,4 = Neither Agree nor Disagree, 3 = Rather Disagree, 2= Disagree, 1 = Strongly Disagree} & \textbf{DiveMate (Mean ± SEM)} \\
\midrule
I had full control over what I was doing. & $6.6 \pm 0.3$ \\
† The movements were the robot's movements, not my own. & $2.4 \pm 0.7$ \\
† My movements just happened, without my intention. & $1.3 \pm 0.3$ \\
I moved the way I wanted. & $6.7 \pm 0.3$ \\
The decision of when and how to move was in my hands. & $6.6 \pm 0.3$ \\
† It felt like the movements were coming from the robot. & $1.6 \pm 0.3$ \\
My movements felt like they were coming from me. & $6.4 \pm 0.3$ \\
† My movements have felt foreign to me. & $1.3 \pm 0.3$ \\
Even if the robot supported me, it was my own movements. & $6.6 \pm 0.4$ \\
I am fully responsible for the movements I have made. & $6.6 \pm 0.3$ \\
\midrule
\textbf{Average Score} & \textbf{$6.5 \pm 0.3$} \\
\bottomrule
\end{tabular}
\\
\vspace{0.5em}
\setlength{\emergencystretch}{3em} % 允许更大的间距以改善对齐
\raggedright
This survey was adapted from the Sense of Agency Scale, which is a measure of consciously perceived control over one's mind, body, and the immediate environment \cite{ZhangXH2024NMI,Tapal2017FroPsy}. Participants ($n = 7$) completed the survey after completing all diving experiments.
\end{table}

%%%%%%%%%%%%%%%%%%%%%%%%%%%%%%%%%%%10pt
\begin{table}[htbp]
\centering
\caption{Comfort and usability survey results for exoskeleton participants}
\label{tab:comfort_usability_combined}
\fontsize{8}{10}\selectfont % 9pt字体，11pt行距
\begin{tabular}{
>{\raggedright\arraybackslash}m{0.395\textwidth} @{}
    >{\centering\arraybackslash}m{0.075\textwidth}   @{}}
\toprule
\raggedright{\textbf{1) Comfort of the DiveMate exoskeleton}}
{\textbf{Question text (4 = Strongly Agree, 3 = Somewhat Agree, 2 = Neither Agree nor Disagree, 1 = Somewhat Disagree, 0 = Strongly Disagree)}}  & \textbf{DiveMate (Mean ± SEM)} \\
\midrule
My skin is free of abrasions and irritations. & $2.9 \pm 0.4$ \\
My exoskeleton is pain free to wear. & $3.0 \pm 0.3$ \\
My exoskeleton is comfortable throughout usage. & $3.1 \pm 0.3$ \\
My exoskeleton looks good. & $3.9 \pm 0.1$ \\
My diving suit is free of wear and tear from my exoskeleton. & $3.6 \pm 0.3$ \\
It is easy to put on and take off my exoskeleton. & $3.4 \pm 0.2$ \\
My exoskeleton fits well. & $3.3 \pm 0.2$ \\
My exoskeleton is durable. & $3.3 \pm 0.2$ \\
The weight of my exoskeleton is manageable. & $3.4 \pm 0.2$ \\
\midrule
\textbf{Average Score} & \textbf{$29.9 \pm 1.2$} \\
\bottomrule
\end{tabular}
\\
\vspace{0.5em}
\setlength{\emergencystretch}{3em} % 允许更大的间距以改善对齐
\raggedright
This survey was adapted from the Orthotics and Prosthetics Users’ Survey \cite{Heinemann2003ProsthetOI}, which acts as a self-report instrument for evaluating clinically 
useful outcomes of prosthetics and orthotics services. Participants  ($n = 7$) completed the survey after completing all diving experiments.
\\
\vspace{1em}
\centering
\begin{tabular}{   
>{\raggedright\arraybackslash}m{0.395\textwidth} @{}
    >{\centering\arraybackslash}m{0.075\textwidth}   @{}}
\toprule
\raggedright{\textbf{2) Usability}}\\
{\textbf{Question text (†: This item is inversely coded; 4 = Strongly Agree, 3 = Somewhat Agree, 2 = Neither Agree nor Disagree, 1 = Somewhat Disagree, 0 = Strongly Disagree)}}  & \textbf{DiveMate (Mean ± SEM)} \\
\midrule
I think that I would like to use this system frequently. & $2.9 \pm 0.3$ \\
† I found the system unnecessarily complex. & $0.9 \pm 0.3$ \\
I thought the system was easy to use. & $3.4 \pm 0.3$ \\
† I think that I would need the support of a technical person to be able to use this system. & $1.1 \pm 0.5$ \\
I found the various functions in this system were well integrated. & $3.4 \pm 0.2$ \\
† I thought there was too much inconsistency in this system. & $0.4 \pm 0.2$ \\
I would imagine that most people would learn to use this system very quickly. & $3.3 \pm 0.4$ \\
† I found the system very cumbersome to use. & $1.0 \pm 0.4$ \\
I felt very confident using the system. & $3.6 \pm 0.3$ \\
† I needed to learn a lot of things before I could get going with this system. & $1.6 \pm 0.5$ \\
\midrule
\textbf{Total usability score (out of 100)} & \textbf{$78.9 \pm 6.5$} \\
\bottomrule
\end{tabular}
\\
\vspace{0.5em}
\setlength{\emergencystretch}{3em} % 允许更大的间距以改善对齐
\raggedright
The System Usability Scale \cite{Jordan1996CRCpress}, which uses a Likert scale, was used to evaluate the usability of the DiveMate exoskeleton. Participants ($n = 7$) completed these surveys after completing all diving experiments.
\end{table}

\subsection{User surveys}
DiveMate preserved users’ strong sense of agency and was considered an easy-to-use and comfortable diving device. The sense of agency is crucial for wearable robotic devices, and it dramatically influences the acceptance of wearable robotic devices among potential users. Firstly, participants were asked to complete a sense of agency survey to assess their feelings of being in control. Each participant rated their level of agreement with a ten-item questionnaire on a scale ranging from 1 (strongly disagree) to 7 (strongly agree) to assess the sense of agency. Participants reported that a strong sense of agency was preserved when using the DiveMate, and the average score on the questionnaire was 6.5 out of 7 (Table \ref{tab:agency_survey}). In addition, participants were asked to complete a System Usability and Comfort Scale survey to determine the comfort and how easy it was to operate the DiveMate. Each participant rated their level of agreement with the items on a scale ranging from 0 (strongly disagree) to 4 (strongly agree). Participants reported that the exoskeleton had a manageable weight, and it was easy to put on and take off. In particular, the diving suit was free of wear and tear from the DiveMate (Comfort survey in Table \ref{tab:comfort_usability_combined}). In addition, participants reported that the exoskeleton was relatively easy to use, with an overall score of 78.9 (Usability survey in Table \ref{tab:comfort_usability_combined}).

\begin{figure*}[htbp]
    \centering{
    \includegraphics[width=0.95\textwidth]{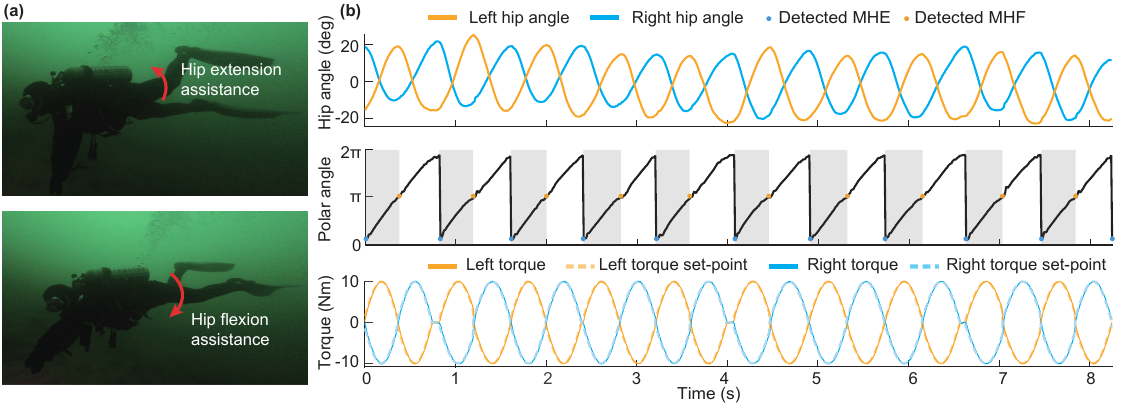}}
    \caption{Open-water validation of field feasibility for the DiveMate system. (a) DiveMate for hip extension assistance during flutter kicking at a depth of 20 meters. (b) Representative time-series data recorded during diving at a depth of 30 meters. Top: Bilateral hip joint angles; Middle: Polar angle trajectory of the hip joint, with detected MHE and MHF; Bottom: Measured bilateral assistive torque output (solid lines) tracked the corresponding torque set-points (dashed lines), confirming stable and accurate torque delivery throughout the dive.}
    \label{Fig_Results_OpenWaterData}
\end{figure*}

\subsection{Open-water validation}
To validate the feasibility of the DiveMate in a realistic diving environment, an open-water trial was conducted at Fuxian Lake, with a surface area of 216 square kilometers, an average depth of 95 m, and a maximum depth of 158 m (Supplementary Video \textcolor{blue}{1}). 
% Fuxian Lake, Yunnan, China, which is one of the deepest freshwater lakes in Asia. 
Two experienced divers participated in the trial, accompanied by two certified diving instructors to ensure compliance with safety protocols. Before the start, the exoskeleton was donned and verified in accordance with standard equipment inspection procedures, confirming its compatibility with the diving suit, tank, and other SCUBA diving equipment.  
Accompanied by instructors, two subjects gradually descended from the surface to a target depth of 30 m. Throughout the descent, the DiveMate remained operational and continuously delivered bidirectional assistive torque to the hip joints. Upon reaching a depth of 30 m, the subjects completed a 100-meter horizontal swim along the lake floor. After the swim, the subjects gradually ascended to the surface, following standard decompression procedures. 
Throughout the entire dive, the DiveMate performed normally, with no mechanical, electrical, or control system malfunctions. The waterproof seal remained intact across the entire depth range, and the assistive torque output remained stable. Fig. \ref{Fig_Results_OpenWaterData} shows the partial hip joint angle trajectories and output torque recorded during the 30-meter diving trial. Both subjects successfully completed the entire dive without any safety incidents and reported that the exoskeleton did not hinder their underwater mobility, buoyancy control, or use of standard diving equipment. 
\section{Discussion and Conclusion}
%总结本研究
This study presents a field-deployable hip exoskeleton designed to address the fundamental challenge of limited endurance in underwater diving. Underwater locomotion imposes higher physiological demands than terrestrial locomotion, with divers expending significant muscular effort and respiratory resources to sustain continuous propulsion. By delivering bilateral assistive torque at the hip joints, DiveMate reduces the muscular burden of underwater locomotion, lowers gas consumption, and ultimately extends the time and distance divers can sustain in the water. 

%为什么选择辅助髋关节
The selection of the hip as the assistance target joint is motivated by biomechanical considerations in both over-ground and underwater environments. During level-ground walking, the ankle joint contributes the largest amount of positive power \cite{Conor2015IJRR}. It is rationally suggested that the ankle should be the most effective assistance target. However, assisting distal joints increases leg inertia and causes extra biological joint torque on the proximal joints, thereby inducing a relatively low energy economy \cite{Conor2015IJRR, Browning2007MSSE}. This explains why hip assistance has achieved the best metabolic reductions in recent on-land exoskeleton studies \cite{SuHao2024Nature}. The case for hip assistance is even stronger underwater, for flutter kicking initiates at the hip and drives the oscillatory motion of the leg, making the hip the primary contributor to thrust generation \cite{Samimy2005SportsEng}. The biomechanical difference between the two environments therefore further emphasizes the importance of the hip in underwater locomotion.

%对耐力的提升
DiveMate has been shown to improve diving endurance, as demonstrated through quantitative assessment involving diving participants. In the diving test, we measured real-time gas pressure changes via our custom pressure gauge and precisely obtained the gas consumption rate. Our exoskeleton reduced CoT by 31.4\% on average compared to diving with the Wo-Assist condition, and thereby the travel distance when consuming the given amount of breathing gas was significantly increased by 42.9\% (Fig. \ref{Fig_Results_Endurance}\textcolor{blue}{d} and \ref{Fig_Results_Endurance}\textcolor{blue}{f}). In terms of dive duration, while the gross gas consumption rate was significantly reduced by 35.5\%, humans can dive longer with an average increment of 54.9\% when consuming a given gas volume (Fig. \ref{Fig_Results_Endurance}\textcolor{blue}{e} and \ref{Fig_Results_Endurance}\textcolor{blue}{g}). Previous exoskeleton studies for terrestrial gait assistance have analyzed the improvement in performance of metabolic rate or CoT under comparative experiments with a given travel distance (treadmill walking or level-ground walking) \cite{Conor2019Science, ZhangXH2024NMI, LiQG2021Science, Collins2022Nature, SuHao2024Nature}. Therefore, the endurance improvement indicated by the travel distance is roughly estimated by the CoT. Here, we design an experimental protocol in which humans consume a given amount of breathing gas, and we observe changes in travel distance. %To the best of our knowledge, this is the first study of diving endurance improvement via exoskeleton technology.

%分析本研究的气体降低效果为什么这么好，水下环境的独特增益机制
%To date, the highest reported net metabolic rate reduction for overground walking with portable exoskeletons stands at 24.3\% \cite{SuHao2024Nature}. In our study, we use the gas consumption rate to estimate human performance, and the net gas consumption rate was reduced by 47.0\% (Fig. \ref{Fig_Results_Endurance}\textcolor{blue}{h}). Although these two results span different measurement modalities (gas consumption vs. indirect calorimetry), environment and task conditions, and cannot be directly compared, the underwater environment offers advantages that partly explain the favorable gas-consumption result. Water buoyancy offsets part of the weight of the body and equipment, making the wearer less sensitive to mass-related energetic loss, and divers must carry lead weights to submerge, whereas DiveMate replaces part of these static loads with an active system that also contributes thrust during kicking, turning static loads into a source of propulsion.

In our study, we use the gas consumption rate to estimate human breathing-gas consumption performance. Although the result cannot be directly compared with terrestrial exoskeleton studies, considering different measurement modalities (gas consumption vs. indirect calorimetry), the underwater environment offers advantages that partly explain the favorable gas-consumption result. Water buoyancy offsets part of the weight of the body and equipment, making the wearer less sensitive to mass-related energetic loss, and divers must carry lead weights to submerge, whereas DiveMate replaces part of these static loads with an active system that also contributes thrust during kicking, turning static loads into a source of propulsion.

%explain why better than other studies:与已有研究对比:1. 助力大小(静态环境) 2.辅助关节
The reduction in gas consumption and muscular effort is consistent with the effects observed in land-based exoskeletons, where robotic assistance has been demonstrated to reduce muscle activation and net metabolic rate \cite{Conor2019Science, Gregg2024SR, LiQG2021Science}. In our study, the net gas consumption rate was significantly reduced by 47\%, after accounting for the resting gas consumption (Fig. \ref{Fig_Results_Endurance}\textcolor{blue}{h}), exceeding the reductions reported by Wang et al. (29.7\%) \cite{WuXY2025TRO} and Wu et al. (20.9\%) \cite{WuXD2025TRO}. Several factors likely contribute to this performance advantage. First, DiveMate delivers a higher peak assistive torque. Based on the experimental parameters reported in \cite{WuXY2025TRO}, the peak assistive torque applied under their experimental conditions was estimated to be approximately 10 Nm, whereas our study used a peak assistance torque of 22 Nm. A higher assistive torque might more effectively substitute for the muscle force required during kicking. It is also worth noting that the static handrail paradigm employed in \cite{WuXY2025TRO} may not allow for a direct comparison with free swimming. Second and most 
fundamentally, the exoskeleton in \cite{WuXD2025TRO} targets the knee joint rather than the hip joint, whereas DiveMate's hip-targeted bilateral assistance addresses the biomechanical bottleneck of flutter kicking more directly.

%大腿幅度和速度都有增加
With the assistance of DiveMate, divers exhibited more powerful kicking, characterized by increased amplitude and velocity (Fig. \ref{Fig_Results_AngleEMG}\textcolor{blue}{d} and \ref{Fig_Results_AngleEMG}\textcolor{blue}{e}). These changes can enhance propulsion efficiency, as a larger kicking amplitude and a higher kicking velocity can generate greater thrust \cite{Goldfarb2024Rbtc}. Although greater thrust is generated by the higher muscle activation, the results of the underwater experiments showed that DiveMate significantly reduced the RMS and peak values of lower-limb muscle activation compared to diving with the Wo-Assist condition (Fig. \ref{Fig_Results_AngleEMG}\textcolor{blue}{a} to \ref{Fig_Results_AngleEMG}\textcolor{blue}{c}). This reduction suggests that divers could maintain or even improve kicking force with lower muscular effort under the DiveMate assistance. Moreover, the actual reduction value in muscle activation attributable to DiveMate is likely underestimated because divers would need to exert higher muscle activation to maintain the same kicking amplitude and velocity without DiveMate assistance.

%机械结构设计总结
DiveMate's portable, symmetrical, and stably attached design is essential to its underwater usability: the modular structure keeps it compatible with existing SCUBA equipment, the symmetric layout helps divers maintain balance under buoyancy, and the combination of a hard structure and soft fabric prevents undesired gaps between the device and the body in wet environments.
Designing the assistance control according to the natural kinematic and muscle activation characteristics of human locomotion is an extensively adopted method, but the corresponding data are not readily accessible for underwater activities. In this study, we developed a hierarchical control framework comprising locomotion intention detection, assistance torque generation, and actuator torque output, based on our previous biomechanical investigation of flutter kicking in diving \cite{yang2026JOB}. The hip-related muscle group activation-inspired assistance torque profile is designed and parameterized. Furthermore, we adopted synchronized bilateral hip assistance with equal torque amplitude to prevent upper-body imbalance from asymmetric bilateral torques.

%控制方法设计总结
%Designing the assistance control according to the natural kinematic and muscle activation characteristics of human locomotion is an extensively adopted method, but the corresponding data are not readily accessible for underwater activities. In this study, we first collected divers' kinematic and muscle activation data and, on this basis, developed a hierarchical control framework comprising locomotion intention detection, assistance torque generation, and actuator torque output. We developed a two-class locomotion intention detection algorithm based on the thigh phase portrait method for continuous and steady assistance \cite{Gregg2017CCTA}. Then, the hip-related muscle group activation-inspired assistance torque profile is designed and parameterized. Furthermore, we adopted synchronized bilateral hip assistance with equal torque amplitude to prevent upper-body imbalance from asymmetric bilateral torques.

%sense of agency
By integrating exoskeleton hardware and assistance control, DiveMate achieves better human–robot coordination. This is reflected in its ability to synchronize with the natural muscle activation patterns and rhythm of flutter kicking, ensuring no extraneous actions are imposed upon voluntary movements. As a result, participants reported a strong sense of agency (Table \ref{tab:agency_survey}), feeling in control rather than passively guided. In addition, participants noted that the exoskeleton was relatively easy to use and comfortable to wear (Table \ref{tab:comfort_usability_combined}), ranking above the 65th percentile of previously surveyed consumer devices \cite{Collins2022Nature, Sauro2011book}. These findings suggest that it is possible to develop wearable systems that are user-friendly, comfortable, and reliable for improving underwater endurance.

%开放水域验证的意义
Unlike pool-based evaluations conducted in calm, shallow, and highly controlled environments, open-water diving at 30 m introduces hydrodynamic variability, thermocline-induced temperature gradients, limited visibility, and the psychological stress of deep open-water exposure. The successful completion of the dive, without reported interference from the exoskeleton on mobility, buoyancy control, or equipment interaction, confirms the field feasibility, waterproofing, equipment compatibility, and operational reliability of DiveMate under the tested conditions.

\section{Limitations}

Several limitations should be considered. First, this single-session study evaluated only the immediate effects of DiveMate in a small, all-male cohort; repeated-session studies with larger and more diverse populations are needed to assess adaptation, cumulative fatigue, and the persistence of the observed benefits. Second, breathing-gas consumption is a practical but indirect indicator of physiological demand and may be affected by respiratory, psychological, thermal, and environmental factors, warranting precise physiological measurements in future studies. Third, because Wo-Assist was not a coupling-matched zero-torque condition, differences in human–robot coupling, mechanical constraints, and local hydrodynamics could not be fully separated from the effects of active assistance; future experiments should therefore include matched zero-torque and no-device controls. Fourth, the assistance profile was not systematically optimized, and human-in-the-loop methods could be used to identify participant-specific and fatigue-adaptive torque amplitude, timing, and profile shape.
Fifth, the controller was evaluated primarily during regular, continuous kicking. Further validation is needed to quantify false triggers, detection latency, and other metrics. Future work should focus on developing a reliable, robust, and low-latency detection algorithm to ensure stable and timely assistance under diverse and dynamically changing underwater conditions. Finally, the evaluation was conducted primarily in a swimming pool, with open-water testing limited to feasibility validation. Extended open-water endurance tests are needed to assess long-term performance and practical applicability.

\section{Conflict of interest}
The authors declare no conflicts of interest.

% \bibliographystyle{ieeetr}
% \bibliography{Sections/Reference20260821}

\printbibliography[title=References]

\end{document}